\documentclass{isprs} % isprs class modified 23-04-2019 (Dennis Wittich)
\usepackage{subfigure}
\usepackage{setspace}
\usepackage{geometry} % added 27-02-2014 Markus Englich
\usepackage{epstopdf}
\usepackage[labelsep=period]{caption}  % added 14-04-2016 Markus Englich - Recommendation by Sebastian Brocks
\usepackage[british]{babel} 
\usepackage[hang]{footmisc}
\usepackage{amsmath}

\DeclareMathOperator*{\argmin}{arg\,min}
\usepackage{amssymb}
\usepackage{booktabs,multirow}
\usepackage{rotating}
\begin{document}

\title{UNDERSTANDING THE IMPACTS OF CROP DIVERSIFICATION IN THE CONTEXT OF CLIMATE CHANGE: A MACHINE LEARNING APPROACH}
\date{}

% KAO: Remove extra spacing
% Anonymous submissions, authors' names should not be visible
\author{
 G. Giannarakis\textsuperscript{1}\thanks{Corresponding author},
 I. Tsoumas\textsuperscript{1,4},
 S. Neophytides\textsuperscript{2},
 C. Papoutsa\textsuperscript{2},
 C. Kontoes\textsuperscript{1},
 D. Hadjimitsis\textsuperscript{2,3}
 }

% KAO: Remove extra newline
% Anonymous submissions, authors' affiliations should not be visible
\address{
	\textsuperscript{1 }BEYOND Centre, IAASARS, National Observatory of Athens,
    Greece - (giannarakis, kontoes, i.tsoumas)@noa.gr\\
	\textsuperscript{2 }ERATOSTHENES Centre of Excellence, Cyprus  - (stelios.neophytides, christiana.papoutsa)@eratosthenes.org.cy\\
    \textsuperscript{3 }Cyprus University of Technology, Cyprus  - (d.hadjimitsis)@cut.ac.cy\\
    \textsuperscript{4 }Wageningen University \& Research, The Netherlands - (ilias.tsoumas)@wur.nl\\
}

% If the corresponding author is NOT the final author, always add a % space before the subsequent comma, i.e.
% first author name\textsuperscript{a,}\thanks{Corresponding author} , % second author name \textsuperscript{b}, etc.
% thanks to Niclas Borlin 05-05-2016

\commission{XX, }{YY} %This field is optional. If filled, XX and YY should be replaced by adequate numbers. See https://www2.isprs.org/commissions/
\workinggroup{XX/YY} %This field is optional.
\icwg{}   %This field is optional.

% KAO: Use times symbol
\abstract{

The concept of sustainable intensification in agriculture necessitates the implementation of management practices that prioritize sustainability without compromising productivity. However, the effects of such practices are known to depend on environmental conditions, and are therefore expected to change as a result of a changing climate. We study the impact of crop diversification on productivity in the context of climate change. We leverage heterogeneous Earth Observation data and contribute a data-driven approach based on causal machine learning for understanding how crop diversification impacts may change in the future. We apply this method to the country of Cyprus throughout a 4-year period. We find that, on average, crop diversification significantly benefited the net primary productivity of crops, increasing it by $2.8\%$. The effect generally synergized well with higher maximum temperatures and lower soil moistures. In a warmer and more drought-prone climate, we conclude that crop diversification exhibits promising adaptation potential and is thus a sensible policy choice with regards to agricultural productivity for present and future.
% These guidelines are provided for the preparation of \textbf{full papers} submitted to ISPRS events (Congress, Geospatial Week, Symposia, smaller events such as workshops). Full papers undergo a double-blind review process. Therefore, the submissions have to be anonymised. If this process leads to acceptance, subsequently a camera-ready manuscript must be submitted following these guidelines (but, of course, incl. author names and affiliation). Depending on the recommendations of the reviewers and the decision of the local programme chair this camera-ready manuscript will be published either in the ISPRS Annals of the Photogrammetry, Remote Sensing and Spatial Information Sciences or The International Archives of the Photogrammetry, Remote Sensing and Spatial Information Sciences, provided it arrives by the due date and corresponds to the guidelines.
% These guidelines are issued to ensure a uniform style for all submitted papers and throughout these two series. To assure timely and efficient production of the Annals and Archives with a consistent and easy to read format, authors must submit their manuscripts in strict conformance with these guidelines. The Society may omit any paper that does not conform to the specified requirements.
}


\maketitle

%\saythanks % added 28-02-2014 Markus Englich

\section{INTRODUCTION}\label{INTRODUCTION}

Globally, agriculture faces the unique challenge of reconciling a growing demand for its products with the substantial pressures stemming from climate change and environmental deterioration. As such, the sustainable intensification of agriculture \cite{tilman2011global} calls for the application of management practices that are sustainable and can adapt to climate change without sacrificing productivity. It encompasses agricultural, social, and environmental dimensions of farming, including its profitability and resilience.

% Of special importance for mitigating climate change is the field of carbon farming, consisting of various agricultural practices that are meant to sequester atmospheric carbon into the soil, and thus make farming more sustainable. 
% However, to achieve the ``sustainable intensification" of agriculture CITE, gains in carbon sequestration should not happen at the expense of crop productivity. It is imperative that productivity effects of such practices are well understood.

As a popular sustainable practice, crop diversification (i.e., cultivating different crops in space and/or time) has been studied extensively, with various positive consequences reported \cite{lin2011resilience}. At the same time, the effects of crop diversification have been found to vary across different environmental conditions \cite{beillouin2021positive}. When analyzing the impacts of crop diversification, it is thus desirable to employ tools that allow for the impacts to depend on relevant environmental variables, such as temperature and soil moisture. An indirect benefit of doing this, is that we may also study questions relating to climate change: will the effects of crop diversification change in a warmer, more drought-prone planet?

With the volume and quality of Earth Observation (EO) data rapidly increasing, data-driven assessments of crop diversification impacts can benefit in terms of scale and reliability \cite{giannarakis2022towards}. Earth Observations are frequently used to engineer large-scale datasets featuring diverse information on agriculture, climate, society and economy \cite{choumos2022towards,drivas2022data}. Artificial Intelligence (AI) techniques including Machine Learning (ML) are used to analyze such datasets, extract insights, and inform stakeholders \cite{sitokonstantinou2023fuzzy,sitokonstantinou2020scalable}. Since impact assessment studies are fundamentally concerned with cause and effect, researchers have been recently using causal machine learning techniques to valorize EO data \cite{jerzak2023integrating,giannarakis2022personalizing,tsoumas2023evaluating,nanushi2022pest} while avoiding the caveats of correlation based ML methods \cite{pearl2009causality}. Even if geospatial data provide ideal information for such analyses, the application of causal ML to tackle questions relating to climate change remains underexplored.

In this work, we propose a causal machine learning approach for investigating the impacts of crop diversification in the context of climate change. In particular, we use geospatial data to train a machine learning model that estimates the effect of crop diversification on productivity as a function of multiple environmental variables. We focus on the relationship of the effects with temperature and soil moisture, and in that way draw insights on the future performance of crop diversification in a warmer, more drought-prone planet.

\section{DATASET AND METHODOLOGY}

We focus on the country of Cyprus from 2019 to 2022 (4-year period). We use the annual MODIS Net Primary Productivity (NPP) product (MOD17A3HGF v006, gridded at 500m) that captures the difference between carbon sequestrated by crops during photosynthesis and carbon released during respiration (in kg C / m$^2$ / year) \cite{running2019mod17a3hgf}. We then use a dataset with all agricultural fields of Cyprus and their crop type from the Cypriot Land Parcel Identification System (LPIS). For each MODIS grid cell and crop type, we compute an annual ``crop abundance" feature by calculating the percentage area of the cell that was covered by the crop type, and then calculate a ``crop diversification" value equal to the number of different crops with non-zero abundance in it. We note that, in that sense, crop diversification is defined on a spatial basis. Similar work can be done by considering crop diversification in a temporal context, akin to crop rotation analyses \cite{gan2015diversifying}.

\begin{figure}[ht!]
\begin{center}
		\includegraphics[width=1\columnwidth]{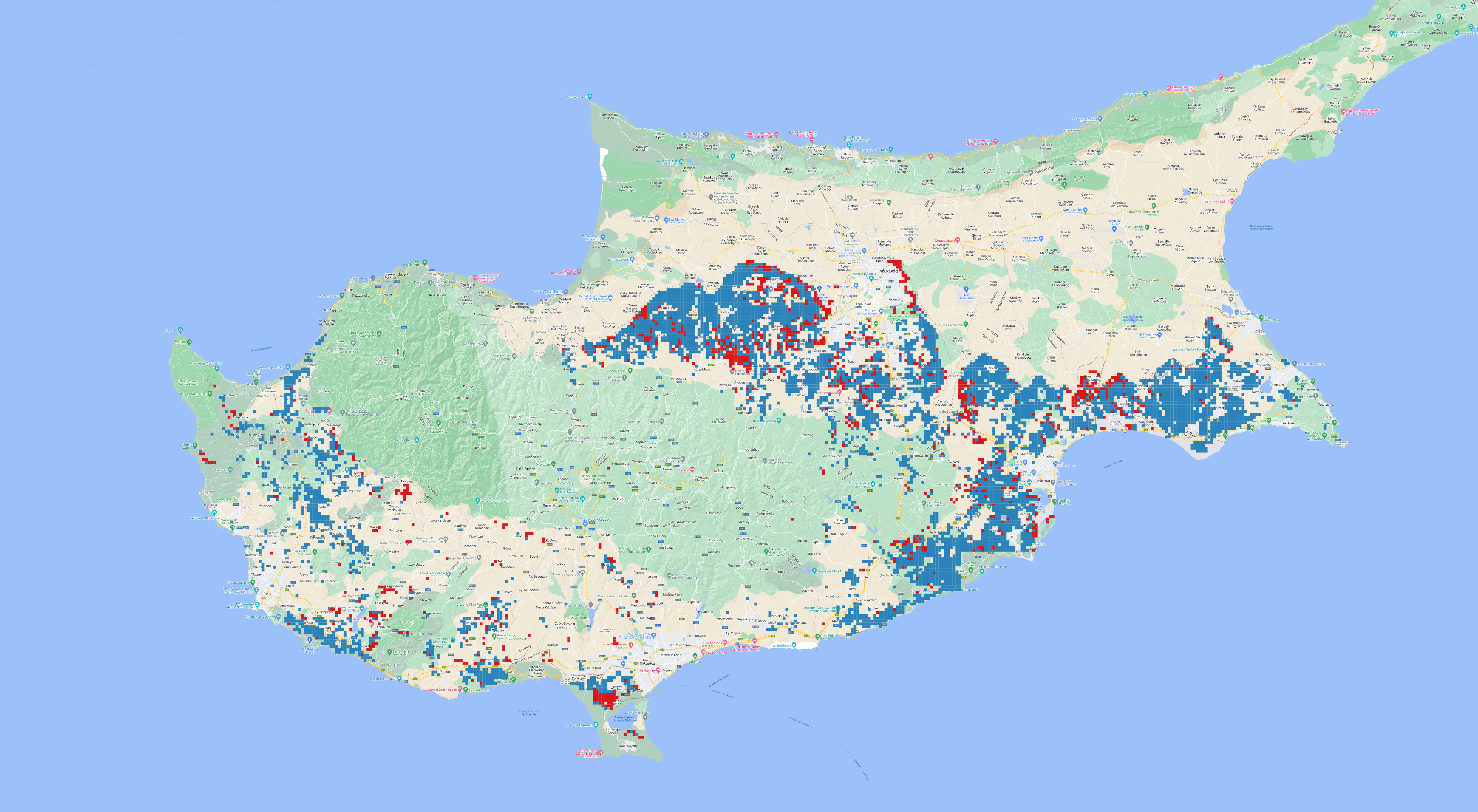}
	\caption{Crop diversification presence (red) and absence (blue) visualized over the main agricultural areas of the country of Cyprus. Both groups are generally found throughout all areas of interest. Moreover, a propensity score filtering is done prior to the analysis to ensure that the groups are comparable.}
\label{fig:cropdiv}
\end{center}
\end{figure}

We join this dataset with climate \cite{abatzoglou2018terraclimate} and soil information \cite{karydas2016modelling} that we downscale to the MODIS grid cell level (see Table \ref{tab:variables}). We finally temporally aggregate all features to obtain an average value for each cell throughout the 4-year study period, and binarize the crop diversification variable at its median value. Figure~\ref{fig:cropdiv} contains the binary crop diversification values over the agricultural areas of Cyprus. Bright grid cells indicate the presence of crop diversification on the final dataset (value greater than median), while dark grid cells indicate the absence of crop diversification (value smaller than median).

\begin{table}[!ht]
\small
\centering
% \resizebox{\columnwidth}{!}{%
\begin{tabular}{lll}
\toprule
\textbf{Id} & \textbf{Variable Description} & \textbf{Unit} \\ \midrule
ws  & Wind speed & m/s                \\
ppt  & Precipitation & mm \\
q & Runoff & mm                  \\
def  & Climate Water Deficit & mm                      \\
srad  & Downward Surface Shortwave Radiation      & W/m$^2$     \\
tmin  & Minimum Temperature           & °C\\
tmax & Maximum Temperature       & °C \\
soilm & Soil Moisture                 & mm \\
soile  & Soil Erosibility           & unitless                  \\
\bottomrule
\end{tabular}%
\caption{Environmental variables used in the study and their unit of measurement.}
\label{tab:variables}
% }
\end{table}

In the context of climate change, we approach the task of comprehending the influence of crop diversification on Net Primary Productivity (NPP) as a Conditional Average Treatment Effect (CATE) estimation task. Using the Potential Outcomes \cite{imbens2015causal} framework, let $Y(T)$ denote the value of a random variable $Y$ if we were to treat a unit with a treatment $T \in \{0, 1\}$. Given a vector of features $X$ describing the units, we want to learn the CATE function:
\begin{equation}
    \theta(x) = \mathbb{E}[Y(1) - Y(0) | X = x]
\end{equation}
We use Double Machine Learning (DML) \cite{chernozhukov2018double} to learn $\theta(x)$ from data, where $T$ is a binary variable for crop diversification, $Y$ is the NPP, and $X$ are the environmental parameters of Table \ref{tab:variables}. DML captures the data generating process using the Partially Linear Model~\cite{robinson1988root}:
\begin{align}
    Y &= \theta(X)\cdot T + g(X) + \varepsilon \label{doublemllinearity}\\
    T &= f(X) + \eta \label{confounding}
\end{align}
where $\theta(X)$ is the CATE, and $g,f$ are arbitrary functions. Notably, \eqref{confounding} monitors confounding as features $X$ affect both the treatment $T$ and outcome $Y$. The CATE $\theta(X)$ is learned though a two-stage estimation process. In the first stage, the outcome $Y$ and treatment $T$ are independently predicted from features $X$, using any ML model. In the second and final stage, $\theta(X)$ is estimated by predicting the residuals of the first model from the residuals of the second model. In the context of the Partially Linear Model shown in \eqref{doublemllinearity} and \eqref{confounding}, this translates to solving the following optimization problem:

\begin{equation}
    \hat{\theta} = \argmin_{\theta \in \Theta}\mathbb{E}\big[(\Tilde{Y} - \theta(X)\cdot \Tilde{T})^2 \big]
    \label{doublemlfinalstage}
\end{equation}
Here, $\Theta$ is the search space of CATE functions, $\Tilde{Y}$ are the residuals of the $Y \sim X$ regression, and $\Tilde{T}$ are the residuals of the $T \sim X$ regression. The linearity assumption imposed by \eqref{doublemllinearity} can be relaxed, allowing for non-parametric CATE estimation.

\begin{figure}[ht!]
\begin{center}
		\includegraphics[width=1\columnwidth]{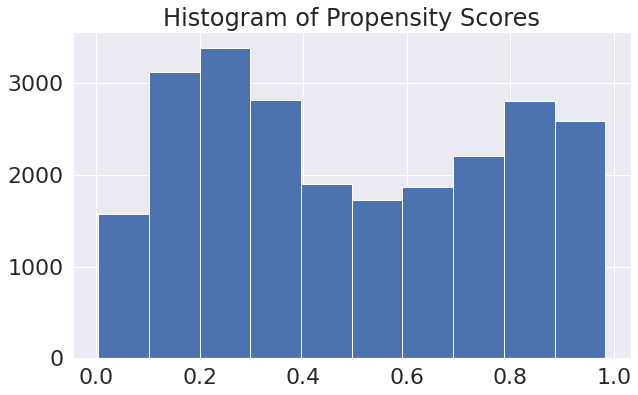}
	\caption{Estimated Propensity scores of the data.}
\label{fig:propensity}
\end{center}
\end{figure}

Before applying Double Machine Learning, we standardize the feature vector $X$ by subtracting the mean and dividing by the standard deviation of each variable. We also fit a Gradient Boosted Propensity Model \cite{chen2020causalml} to estimate the propensity score $\mathbb{P}(T = 1 | X = x)$ of each sample, and filter extreme scores ($<0.2$ or $>0.8$) to aid the overlap between the treatment and control groups \cite{imbens2015causal}. The distribution of propensity scores prior to filtering is shown in Figure~\ref{fig:propensity}. After filtering, there are $14201$ samples left, out of which $6661$ belong to the treated (crop diversification) group and $7540$ are part of the control (no crop diversification) group.

We split the dataset into train ($80\%$) and test ($20\%$) sets to fit the first stage DML models. We carry out a 3-fold cross validation on the train set and separately model $\mathbb{E}[Y | X]$ and $\mathbb{E}[T | X]$. A Random Forest model was selected for both the regression ($Y\sim X$) and classification ($T\sim X$) tasks, outperforming Lasso regression, Logistic regression, and Gradient Boosting models. To diagnose overfitting, we then use the held-out test set to assess the predictive performance ($R^2$ for regression, F-1 score for classification) of the trained models and compare it to the performance of the train test. Table~\ref{tab:fitting} contains the results of the first stage models.

\begin{table}[!ht]
\small
\centering
% \resizebox{\columnwidth}{!}{%
\begin{tabular}{lll}
\toprule
\textbf{Task} & \textbf{Train} & \textbf{Test} \\ \midrule
$\mathbb{E}[Y | X]$  & $0.78 $ & $0.80$                \\
$\mathbb{E}[T | X]$  & $0.59$ & $0.60$ \\
\bottomrule
\end{tabular}%
\caption{Performance ($R^2$ for $\mathbb{E}[Y | X]$, F-1 score for $\mathbb{E}[T | X]$) of the selected first stage Random Forest models. Outcome Y is Net Primary Productivity, (binary) treatment T is crop diversification, X is the vector of features.}
\label{tab:fitting}
% }
\end{table}

\begin{figure*}[ht!]
    \centering
	\includegraphics[width=0.9\linewidth]{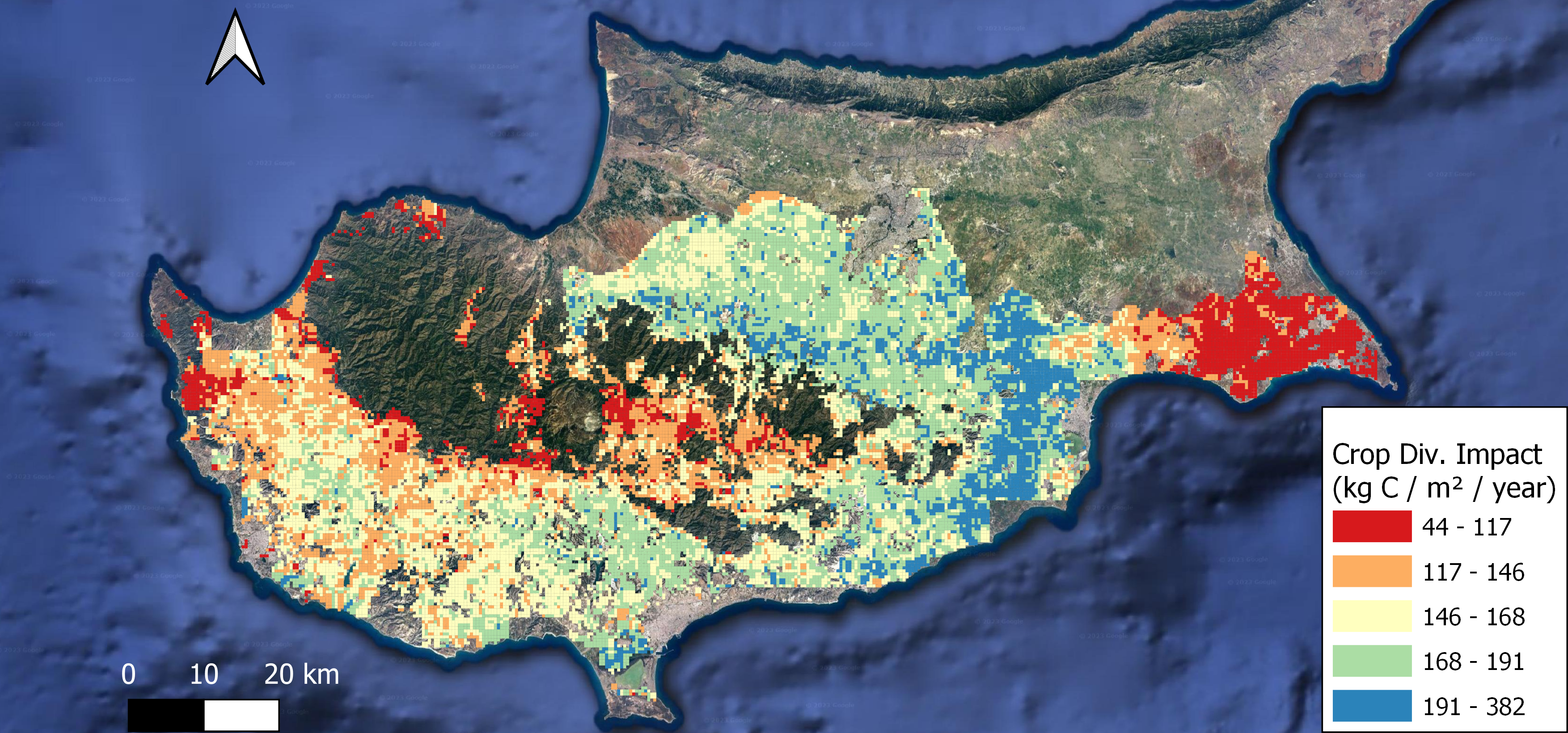}
	\caption{CATE point estimates for crop diversification visualized over all areas of Cyprus for which we had LPIS, environmental, and net primary productivity data. The estimated crop diversification impacts range from 44 to 382 kg C / m$^2$ / year, with an average treatment effect of $157$ kg C / m$^2$ / year. Considerable spatial heterogeneity is found, consolidating the importance of targeted agricultural policy making.}
\label{fig:map}
\end{figure*}

The predictive performance on the train and test sets were comparable for both first stage DML tasks indicating the lack of overfitting. Having selected the Random Forest models through this procedure, we proceed to the final stage of DML, where we predict the residuals of $Y\sim X$ from the residuals of $T\sim X$. To retain model interpretability, we fit the Linear DML model shown in equations \eqref{doublemllinearity} and \eqref{confounding}.

\section{EVALUATION AND RESULTS}

A significant Average Treatment Effect (ATE) of crop diversification on NPP is found (point estimate = $157$, 95\% CI = $[137, 177]$). This indicates that, on average, diversifying crops had a positive impact on their primary productivity (i.e., it increased it by $157$ kg C / m$^2$ / year). Interpreted in the context of the mean net primary productivity (which was equal to $5544$ kg C / m$^2$ / year), this translates to an estimated average increase of $2.8\%$ due to the implementation of a crop diversification practice on a cell over a single year. The effect is both positive and statistically significant (p-value $< 0.001$), corroborating growing evidence on the productivity benefits of the practice \cite{beillouin2019evidence}. 

Figure \ref{fig:map} shows all CATE estimates (i.e., expected crop diversification impacts) provided by the trained model. As expected, spatial proximity coincides with similar crop diversification impacts \cite{tobler1970computer}. The geospatial information featured in Figure~\ref{fig:map} can be exploited for targeted agricultural policy making, e.g. by incentivizing the implementation of crop diversification on the basis of its predicted impact.

\begin{figure}[ht!] 
\begin{center}
		\includegraphics[width=0.93\columnwidth]{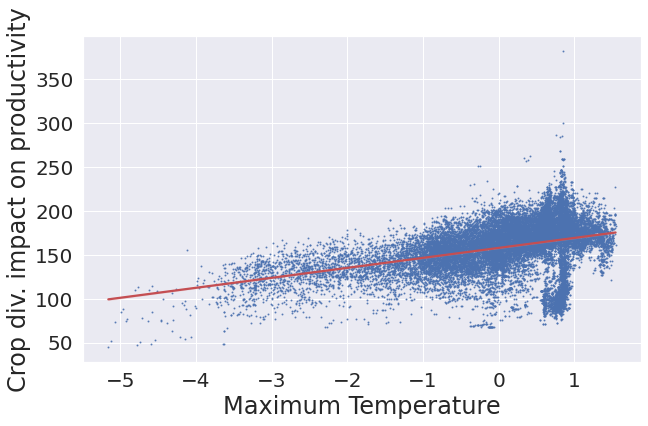}
	\caption{The relationship between crop diversification impacts on net primary productivity and maximum temperature.}
\label{fig:temp-cate}
\end{center}
\end{figure}

\begin{figure}[ht!] 
\begin{center}
		\includegraphics[width=0.93\columnwidth]{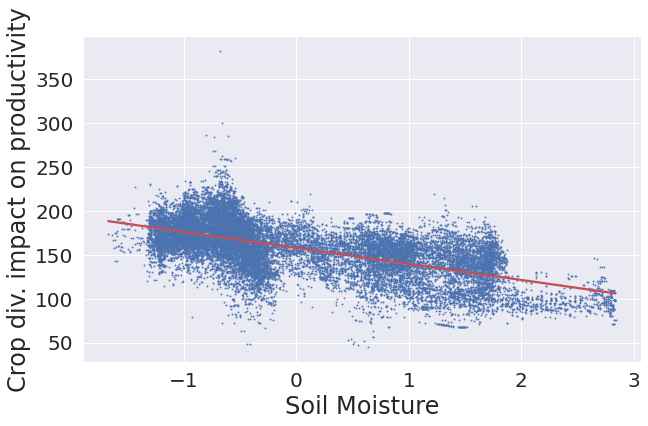}
	\caption{The relationship between crop diversification impacts on net primary productivity and soil moisture.}
\label{fig:soil-cate}
\end{center}
\end{figure}

The relationship between maximum temperature, soil moisture and crop diversification impacts on NPP is seen in Figures \ref{fig:temp-cate} and \ref{fig:soil-cate}. We observe a linear trend for both relationships, albeit of opposite slope; the impact of diversification on productivity appears to increase with temperature and decrease with soil moisture. Because both maximum temperature and soil moisture variables are standardized, the horizontal axis unit is the number of standard deviations away from the variable mean. Table~\ref{tab:clima} features these statistics for both variables.

\begin{table}[!ht]
\small
\centering
% \resizebox{\columnwidth}{!}{%
\begin{tabular}{llll}
\toprule
\textbf{Variable} & \textbf{Mean} & \textbf{Std. Deviation} & \textbf{Unit} \\ \midrule
Maximum temperature  & $25.2 $ & $1.7$ &   °C             \\
Soil moisture  & $45.9$ & $9.2$  & mm\\
\bottomrule
\end{tabular}%
\caption{Observed mean and standard deviation of maximum temperature and soil moisture variables, averaged over 2019-2022 and over all grid cells of Cyprus.}
\label{tab:clima}
% }
\end{table}

Examining Figure~\ref{fig:temp-cate}, we are therefore able to study crop diversification impacts in the context of climate change. Due to the increasing trend found, we can infer that an increase in maximum temperature by $1.7$°C (i.e., by $1$ standard deviation) will generally lead to higher crop diversification impacts. We note that an increase of this magnitude is associated with low GHG emissions scenarios (RCP1.9, RCP2.6). For higher emission, less optimistic scenarios that lead to greater maximum temperature values the extrapolation needed is larger. In this case, while the overall outlook for future crop diversification impacts remains positive, more uncertainty is introduced.

Similar insights are derived from Figure~\ref{fig:soil-cate}, by interpreting the relation between soil moisture and crop diversification impacts in climate change terms. An increase in maximum temperature will lead to dryer soils that are associated with higher crop diversification impacts on productivity. The decreasing trend of crop diversification impacts as a function of soil moisture is stable for more than 1 standard deviation to the left of the mean. This remark provides confidence on the robustness of crop diversification (with regards to productivity) against dryer environmental conditions that may be encountered in the future.

\begin{figure}[ht!] 
\begin{center}
		\includegraphics[width=1\columnwidth]{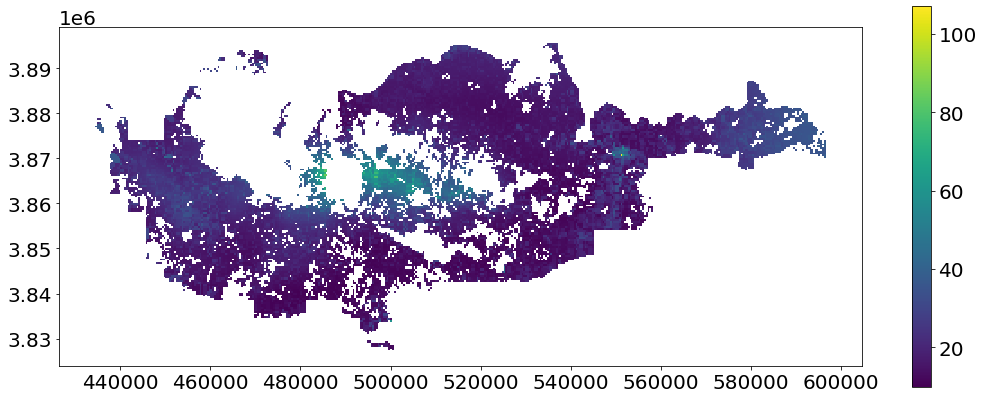}
	\caption{Standard error map for each crop diversification CATE estimate, measured in kg C / m$^2$ / year. Standard errors are generally uniform throughout the map. Higher values are concentrated at the centre of the country that do not correspond to agricultural areas (see Figure \ref{fig:agri-areas}).}
\label{fig:stderr}
\end{center}
\end{figure}

The DML model also provides uncertainty estimates on the grid cell level CATE predictions of crop diversification impacts in the form of standard errors. Figure~\ref{fig:stderr} shows the standard error that corresponds to each CATE estimate. We remark that compared to the average crop diversification effect of $157$ kg C / m$^2$ / year, standard error values are small, ranging from $15$ to $50$ kg C / m$^2$ / year for most areas. This translates to statistically significant individual CATE estimates, with $99.9\%$ of the grid cells having a p-value $<0.05$.

\begin{figure}[ht!] 
\begin{center}
		\includegraphics[width=1\columnwidth]{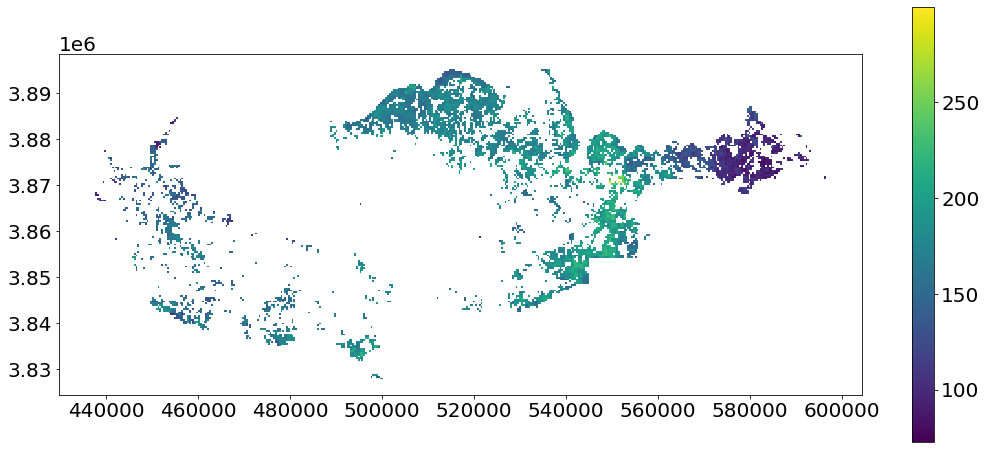}
	\caption{Map showing the estimated crop diversification impacts over the agricultural areas of Cyprus. Each grid cell area is covered by at least 50\% agricultural land, as declared in the official Land Parcel Identification System (LPIS).}
\label{fig:agri-areas}
\end{center}
\end{figure}

It should be noted that for visualization purposes, we so far chose to visualize CATE results throughout the entire Cyprus area over which we had data (Figures~\ref{fig:map} and ~\ref{fig:stderr}). For sensible agricultural policy making, we should exclusively focus on agricultural areas and derive insights based on them. Here we define ``agricultural areas" as the grid cells whose area is covered by agricultural parcels by at least $50\%$. Figure~\ref{fig:agri-areas} visualizes those areas, and reports the corresponding crop diversification impacts over them. The distribution of the estimated impacts does not differ significantly from the one reported across the entire country in Figure~\ref{fig:map}.

\begin{figure*}[ht!]
    \centering
	\includegraphics[width=1\linewidth]{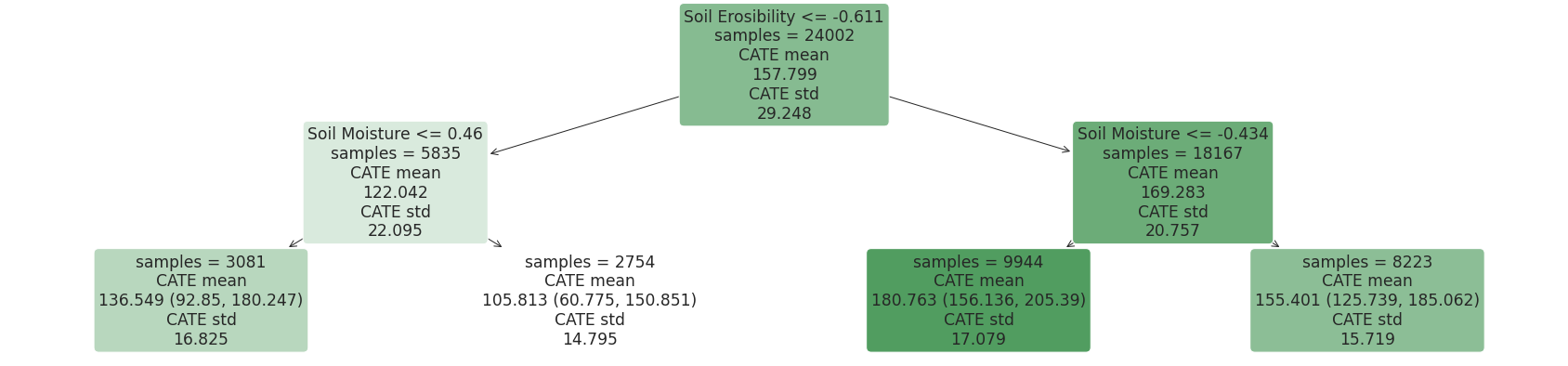}
	\caption{Unveiling the environmental drivers of crop diversification impacts with a tree interpreter. To be read from top to bottom, going left if the Boolean condition at the top of each box is true, and right if it is false. The tree reports the sample size of each leaf, its mean crop diversification impact, standard deviation and $95\%$ confidence intervals on the mean.}
\label{fig:tree}
\end{figure*}

The trained Double Machine Learning model can also be interpreted using explainable AI (XAI) methods in order to reveal the conditions that determine the magnitude of crop diversification impacts. Here, we are using a tree interpreter \cite{battocchi2019econml} to reveal the environmental factors (Table~\ref{tab:variables}) that are driving the estimated effect magnitudes (Figure~\ref{fig:tree}). Soil erosibility is found to be the most influential environmental parameter for crop diversification impacts, followed by soil moisture. In particular, higher soil erosibility conditions, combined with lower soil moisture lead to the highest diversification productivity impacts, while lower soil erosibility combined with high soil moisture lead to lower crop diversification impacts.

% \begin{figure}[ht!] 
% \begin{center}
% 		\includegraphics[width=1\textheight,angle=90]{tree.png}
% 	\caption{The relationship between crop diversification impacts on net primary productivity and soil moisture.}
% \label{fig:soil-cate}
% \end{center}
% \end{figure}

% \begin{sidewaysfigure}
%     \centering
%     \includegraphics[width=0.9\textheight,height=0.3\textwidth]{tree.png}
%     \caption{Intensity measure correlation}
%     \label{fig:IMs matrix correlation}
% \end{sidewaysfigure}

\section{CONCLUSIONS}

This study investigated the effect of diversifying crops in space on the net primary productivity of land. We combined satellite remote sensing data with geo-referenced crop type maps to engineer a heterogeneous Earth Observation dataset. We then trained a Double Machine Learning model that is able to estimate the impacts of crop diversification on productivity, as a function of environmental conditions. The trained model provided robust uncertainty estimates, and was analyzed with an explainable AI method to unveil the environmental drivers of crop diversification performance. 

A statistically significant Average Treatment Effect of crop diversification on net primary productivity was found, that was further modified by both maximum temperature and soil moisture. Results were robust from the statistical perspective, as indicated by small standard errors of estimates, and by the overwhelming majority of estimates ($99.9\%$) having a p-value smaller than $0.05$.

Overall, our work contributes to the growing literature on the effects of crop diversification across multiple dimensions \cite{feliciano2019review,louhichi2017does,kumar2019impact}. While the average effect of crop diversification on NPP is found to be both positive and statistically significant, we note that analogous works in other areas of the world do not report similar productivity gains due to diversification, either of crops \cite{giannarakis2022towards} or of biodiversity \cite{dee2023clarifying}. In the context of the European Union's Common Agricultural Policy, this remark highlights the importance of targeted policy making, as a crop diversification measure might be significantly positive for ecosystem productivity in one area, and fail to deliver any tangible benefits in another.

Importantly, our approach also allows for the interpretation of crop diversification in the context of a changing climate. From the perspective of climate change adaptation in particular, the impact of crop diversification on productivity appears to benefit from the imminent increase in maximum temperature and decrease in soil moisture. We thus conclude that encouraging the diversification of crops in Cyprus is a sensible policy choice as far as productivity is concerned, for both present and future. Enriching the analysis with other environmental, social, and economic parameters to obtain a more holistic view of crop diversification impacts towards the sustainable intensification of agriculture is future work.

\section{ACKNOWLEDGEMENTS}

We acknowledge the ``EXCELSIOR": ERATOSTHENES: Excellence Research Centre for Earth Surveillance and Space Based Monitoring of the Environment H2020 Widespread Teaming project (www.excelsior2020.eu). The ``EXCELSIOR" project has received funding from the European Union’s Horizon 2020 research and innovation programme under Grant Agreement No 857510, from the Government of the Republic of Cyprus through the Directorate General for the European Programmes, Coordination and Development and the Cyprus University of Technology. We thank the Cyprus Agricultural Payments Organization (CAPO) for providing the LPIS data.

{
	\begin{spacing}{1.17}
		\normalsize
		\bibliography{ISPRSguidelines_authors} % Include your own bibliography (*.bib), style is given in isprs.cls
	\end{spacing}
}

\end{document}